\documentclass[10pt,twocolumn,letterpaper]{article}
\usepackage[T1]{fontenc}
\usepackage{authblk}
\usepackage[T1]{fontenc}
\usepackage[utf8]{inputenc}

\usepackage{cvpr}              

\usepackage{graphicx}
\def\shrug{\texttt{\raisebox{0.75em}{\char`\_}\char`\\\char`\_\kern-0.5ex(\kern-0.25ex\raisebox{0.25ex}{\rotatebox{45}{\raisebox{-.75ex}"\kern-1.5ex\rotatebox{-90})}}\kern-0.5ex)\kern-0.5ex\char`\_/\raisebox{0.75em}{\char`\_}}}
\newcommand{\idk}{SHRUG-FM} 

\usepackage{verbatim}

\usepackage{lipsum}

\usepackage{tcolorbox}
\usepackage{xcolor}

\usepackage{tcolorbox}
\usepackage{xcolor}

\newcommand{\scorebox}[1]{%
\begin{tcolorbox}[
colback=red!4,
colframe=red!60!black,
boxrule=0.6pt,
arc=4pt,
left=6pt,right=6pt,top=6pt,bottom=6pt
]
#1
\end{tcolorbox}
}

\definecolor{cvprblue}{rgb}{0.21,0.49,0.74}
\usepackage[pagebackref,breaklinks,colorlinks,allcolors=cvprblue]{hyperref}

\def\paperID{50} 
\def\confName{EarthVision}
\def\confYear{2026}

\title{SHRUG-FM: Reliability-Aware Foundation Models for Earth Observation}

\makeatletter \renewcommand\AB@authnote[1]{} \makeatother

\author{Maria Gonzalez-Calabuig\thanks{Contributed equally}$^{1}$ \; Kai-Hendrik Cohrs$^{*1}$ \; Vishal Nedungadi$^{*2}$ \quad Zuzanna Osika$^{3}$ \; \\ Ruben Cartuyvels$^{4}$ \quad Steffen Knoblauch$^{5}$ \quad Joppe Massant$^{6}$ \quad \\Shruti Nath$^{7}$ \quad Patrick Ebel$^{8}$ \quad Vasileios Sitokonstantinou$^{2}$}

\affil[ ]{$^{1}$Universitat de València, $^{2}$Wageningen University \& Research, $^{3}$Delft University of Technology, $^{4}$European Space Agency, \quad  $^{5}$Heidelberg University, $^{6}$Ghent University, $^{7}$University of Oxford, $^{8}$Google Research}

\affil[ ]{\texttt{\{kai.cohrs, maria.gonzalez-calabuig\}@uv.es}}
\affil[ ]{\texttt{\{vishal.nedungadi, vassilis.sitokonstantinou\}@wur.nl, z.osika@tudelft.nl}}
\affil[ ]{\texttt{ruben.cartuyvels@esa.int, steffen.knoblauch@uni-heidelberg.de, joppe.massant@ugent.be}}
\affil[ ]{\texttt{shruti.nath@physics.ox.ac.uk, pwjebel@google.com}}

\begin{document}
\twocolumn[{%
\renewcommand\twocolumn[1][]{#1}%
\maketitle
\centering
\includegraphics[width=0.85\linewidth]{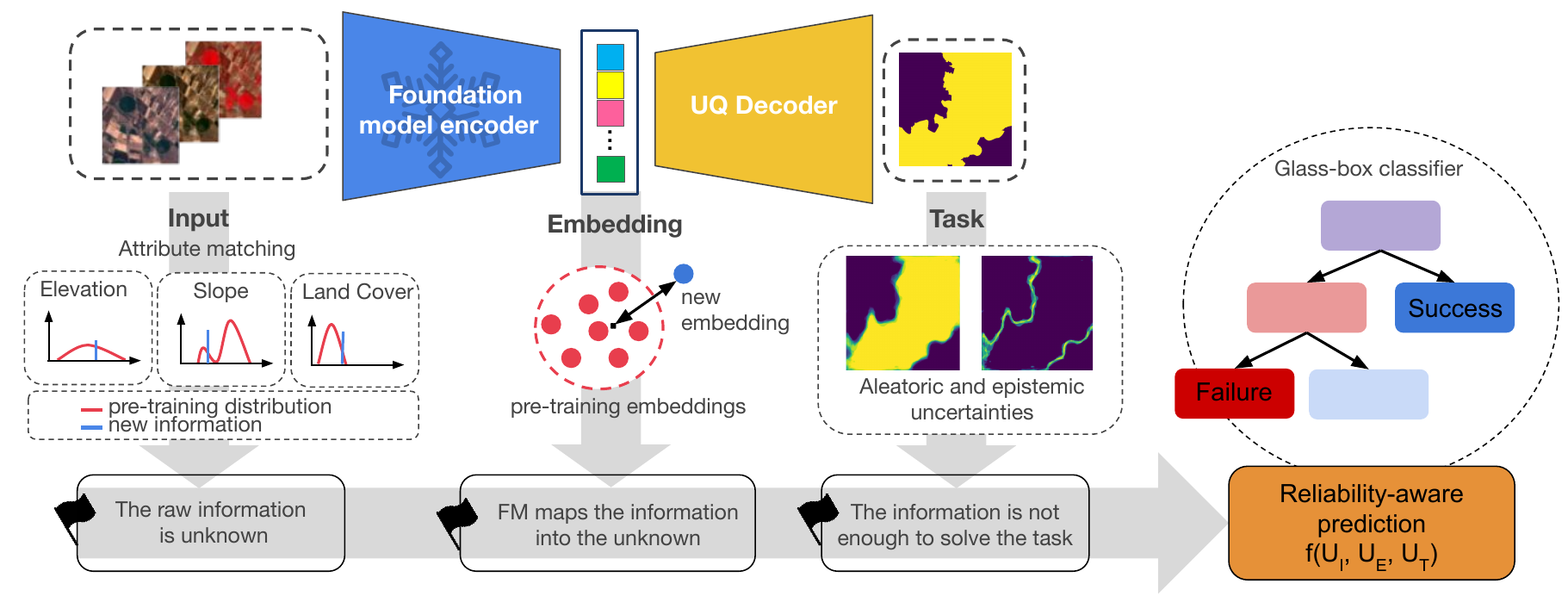}
\captionof{figure}{\textbf{The \idk{} framework.}  It combines three complementary signals: (i) OOD at the input ($\mathcal{U}_{I}$) through geophysical features, (ii) OOD detection in the embedding space ($\mathcal{U}_{E}$) through clustering and (iii) task-specific predictive uncertainty ($\mathcal{U}_{T}$). These are combined to enable reliability-aware predictions, flagging or abstaining from low-confidence outputs
\label{fig:pipeline}
}
\vspace{2.5em}
}]
\insert\footins{\noindent\footnotesize$^*$Equally contributing first authors}

\begin{abstract}
Geospatial foundation models (GFMs) for Earth observation often fail to perform reliably in environments underrepresented during pretraining. We introduce \idk{}, a framework for reliability-aware prediction that enables GFMs to identify and abstain from likely failures. Our approach integrates three complementary signals: geophysical out-of-distribution (OOD) detection in the input space, OOD detection in the embedding space, and task-specific predictive uncertainty. We evaluate \idk{} across three high-stakes rapid-mapping tasks: burn scar segmentation, flood mapping, and landslide detection. Our results show that \idk{} consistently reduces prediction risk on retained samples, outperforming established single-signal baselines like predictive entropy. Crucially, by utilizing a shallow "glass-box" decision tree for signal fusion, \idk{} provides interpretable abstention thresholds. It builds a pathway toward safer and more interpretable deployment of GFMs in climate-sensitive applications, bridging the gap between benchmark performance and real-world reliability. Our source code is available at \url{https://github.com/vishalned/SHRUG-FM}.
\end{abstract}
\section{Introduction}
\label{sec:intro}

Following the success of foundation models (FMs) in natural language processing and computer vision, geospatial foundation models (GFMs) for Earth observation (EO) are gaining traction~\citep{xiao2025foundationmodelsremotesensing}. Trained on large-scale datasets gathered by satellites such as the Sentinel and Landsat missions, these models aim to learn transferable representations of Earth's surface. Recent examples, ranging from early milestones like Clay~\citep{clayWebsite}, SSL4EO-S12~\citep{wang2023ssl4eo}, and Scale-MAE~\citep{reed2023scale} to more recent multimodal and multitemporal architectures like SkySense~\citep{guo2024skysense}, Prithvi-EO-2.0~\citep{szwarcman2025prithvi}, TerraMind~\citep{jakubik2025terramind}, AlphaEarth~\cite{brown2025alphaearth}, and TESSERA~\citep{feng2025tessera}, capture spatial and spectral patterns across diverse regions and timescales. These general-purpose representations are provided as precomputed embeddings or pre-trained models and can be applied to downstream tasks, out-of-the-box or with minimal finetuning to derive application-specific features.

However, ongoing research questions the reliability of FMs for EO in real-world scenarios~\citep{zhao2025exebenchbenchmarkingfoundationmodels,marsocci2025pangaeaglobalinclusivebenchmark}. These models can fail under spatial extremes, such as underrepresented geographies (e.g., deserts, polar regions, high latitudes) and under temporal extremes, such as extreme weather events or long-term environmental shifts like droughts, due to gaps in geographical or seasonal coverage of the data used for pretraining~\cite{wang2023ssl4eo,zhao2025exebenchbenchmarkingfoundationmodels}. 
Most models lack built-in mechanisms to detect out-of-distribution (OOD) inputs or quantify uncertainty, often leading to overconfident predictions \cite{guo2017calibration, hendrycks2016baseline, gawlikowski2023survey}. Recent benchmarks such as REOBench~\citep{li2025reobenchbenchmarkingrobustnessearth}, GeoBench~\citep{lacoste2023geobenchfoundationmodelsearth} and PANGAEA~\citep{marsocci2025pangaeaglobalinclusivebenchmark} evaluate GFMs under conditions that better reflect real-world deployment, including shifts in geography, seasonality and sensor type. The results show that GFMs are regularly outperformed by simple supervised baselines, highlighting the need for systematic reliability assessments of their predictions, including OOD detection and uncertainty estimation.

To address this, we present \idk{}, 
a framework for reliability-aware prediction in GFMs. Our core intuition is that model failures stem from two primary sources: (a) a divergence between the data the model was pre-trained on and the data it encounters during deployment (i.e., distribution shift), and (b) high predictive uncertainty inherent to the downstream task model. Rather than forcing a GFM to predict blindly, \idk{} acts as a comprehensive selective prediction mechanism spanning the entire modeling pipeline (Fig. \ref{fig:pipeline}). It integrates three complementary signals: (1) OOD detection in the input space ($\mathcal{U}_{I}$), which leverages interpretable geophysical features (e.g., elevation, land cover) to signal unfamiliar physical environments, (2) OOD detection in the foundation model's embedding space ($\mathcal{U}_{E}$) to signal unfamiliar representations, and (3) downstream task-specific uncertainty quantification ($\mathcal{U}_{T}$) derived from the downstream decoder. We formalize a selective predictor $h$ equipped with a selection function $g$ that learns when to accept a prediction and when to abstain based on the fusion of these three signals.

We demonstrate \idk{} across multiple challenging downstream EO tasks that are sensitive to real-world distribution shifts: burn scar segmentation on ExEBench~\citep{zhao2025exebenchbenchmarkingfoundationmodels}, flood mapping using the WorldFloods dataset~\citep{PortalesJulia2023}, and landslide detection using the Landslide Reference Data dataset~\citep{Orynbaikyzy2025}. Using models trained on the SSL4EO-S12 dataset~\citep{wang2023ssl4eo}, our empirical results show that \idk{} successfully identifies and discards unreliable predictions across diverse geographies and disaster events, effectively balancing benchmark performance and real-world deployment safety.
 

In summary, our main contributions are as follows:
\begin{itemize}
    \item \textbf{We formalize and quantify distinct sources of failure in GFMs.} We identify three complementary reliability signals in GFM deployment pipelines, i.e., input-level ($\mathcal{U}_{I}$), embedding-level ($\mathcal{U}_{E}$), and task-level ($\mathcal{U}_{T}$). 
    \item \textbf{We introduce a unified selective prediction framework for GFMs.} We design \idk{}, a model-agnostic framework built around a selective classifier that integrates these reliability signals into a single interpretable abstention mechanism.
    \item \textbf{We empirically evaluate reliability:} Across burn scar segmentation, flood mapping, and landslide detection tasks, we show that combining input-, embedding-, and task-level signals consistently reduces prediction risk on retained samples.
\end{itemize}


\section{Related Work}

The literature on GFMs is growing at an increasing pace~\citep{clayWebsite,reed2023scale,guo2024skysense,szwarcman2025prithvi,jakubik2025terramind,brown2025alphaearth,feng2025tessera}, but uncertainty and reliability-aware design are thus far barely accounted for in prior work.
The first step in building general-purpose GFMs is to curate extensive, global pre-training datasets, such as the SSL4EO-S12 dataset \cite{wang2023ssl4eo} used in our experiments.
Nonetheless, recent evaluation studies have established that GFM performance breaks down in spatiotemporal generalization settings and OOD scenarios~\cite{lacoste2023geobenchfoundationmodelsearth,marsocci2025pangaeaglobalinclusivebenchmark,zhao2025exebenchbenchmarkingfoundationmodels,li2025reobenchbenchmarkingrobustnessearth}. The nature of why or when GFM generalization breaks down is not clearly understood and established models lack flagging mechanisms to facilitate user-centric diagnostics. Given the relevance of GFMs for safety-critical decision-making, developing a framework to increase trust and reliability is critical for safe GFM deployment \cite{dramsch2025explainability}, which is the main objective of this work.

Prior works \cite{dionelis2024fine,dionelis2025care,kondylatos2025generalizationrepresentationuncertaintyearth,rey2025uncertainty} estimate predictive uncertainty and evaluate selective prediction.
Here, we complement predictive uncertainty estimates with input- and latent-space OOD signals and show that their combination jointly improves selective prediction accuracy. For input- and latent-space diagnostics, we draw inspiration from the broader OOD literature by \citet{yousefzadeh2022decision,yousefzadeh2023ambiguity} to define reliability flags based on distances between sample clusters in the latent space.
Similarly, \citet{ekim2025distribution} develops a method for OOD detection for EO data in latent space, but does not evaluate selective prediction nor the integration with uncertainties. Our task-based uncertainty signal is based on ensemble estimates as in \cite{Lakshminarayanan2017simple}. We ground our OOD detection in physical characteristics by including hydro-environmental attribute databases, as in \cite{nearing2024global}, to facilitate interpretability and diagnostics.

To conclude, prior to our efforts, no single contribution in the EO domain integrated input- and latent-space distribution analysis with predictive uncertainty into a unified selective prediction system. Closing this gap in the literature by proposing a combined reliability and confidence flagging framework is the core contribution of this work.

\section{Methodology}

Our proposed framework, \idk{}, is built on top of any uncertainty-aware model as a selective prediction mechanism (see Fig. \ref{fig:pipeline}). Specifically, we define a selective predictor $h$ that combines three signals extracted from different stages of the uncertainty-aware model $f$, together with a classifier $g$ that determines whether to accept or abstain from a prediction. Given a feature space $\mathcal{X}$ and a set of labels $\mathcal{Y}$, an uncertainty-aware model $f$ is defined as a function of  $f: \mathcal{X} \xrightarrow{} \mathcal{Y}$, trained on top of a foundation model encoder. The model's output can be decomposed in two components $f(x) = (f_T(x), f_{UQ}(x))$, where $f_T(x)$ denotes the task-specific prediction (e.g., a binary classification output), and $f_{UQ}(x)$ represents the associated uncertainty quantification metrics. Formally, the selective predictor can be expressed as:
\begin{equation}
    h(x,\mathcal{U}_I, \mathcal{U}_E) = \begin{cases}
    f_T(x) & \text{if $g(\mathcal{U}_I, \mathcal{U}_E, f_{UQ}(x)) = 1$} \\
    \scriptsize\shrug & \text{if $g(\mathcal{U}_I, \mathcal{U}_E, f_{UQ}(x)) = 0$}
  \end{cases}
\end{equation}

\noindent
where {\scriptsize\shrug } is a special label indicating the absence of prediction and $\mathcal{U}_I$, $\mathcal{U}_E$, and $f_{UQ}(x) = \mathcal{U}_T$ are the signals associated with the input, the embeddings, and the task, respectively. In the context of remote sensing (RS) downstream applications, this would signal to the user to inspect the prediction by hand before making safety-critical decisions. 

\paragraph{Selective prediction objective.}
Let $y_\text{fail}(x)\in\{0,1\}$ be the task-dependent failure indicator (for example $y_\text{fail}(x)=\mathbf{1}\{\text{F1}(x)<\tau\}$ with threshold $\tau$). The selection function $g:\mathcal{X}\to\{0,1\}$ accepts a prediction if and only if $g(x)=1$.
We define coverage and selective risk as
$$
\mathrm{Coverage}(g)=\mathbb{E}[g(x)],\,
\mathrm{Risk}(g)=\mathbb{E}\big[\,y_\text{fail}(x)\mid g(x)=1\,\big].
$$
In this work, we evaluate selection functions by the standard risk–coverage trade-off (and its area summarization), i.e., the goal is to minimize risk at fixed coverage (or equivalently to minimize AURC).

\paragraph{Why three signals?}
Failures in downstream tasks can arise for different reasons: (i) the input corresponds to under-represented physical conditions and geographies (input-level OOD), (ii) the foundation model's predicted representation is unfamiliar (embedding-level OOD), or (iii) the downstream model is simply uncertain about the label even when inputs look reasonable (task uncertainty). These failure modes are distinct in practice and in detection power, motivating a fused selective mechanism.

\paragraph{$\mathcal{U}_I$ -- The input signal}
\begin{quote}
\emph{I’ve never seen anything like this before!}\\ 
\hspace*{\fill} - \idk{}
\end{quote}

GFMs are trained on vast datasets with scene counts ranging from hundreds of thousands to billions~\cite{marsocci2025pangaeaglobalinclusivebenchmark}. Some sample the data only at a specialized location, and others cover the full globe~\cite{jakubik2023foundationmodelsgeneralistgeospatial, szwarcman2025prithvi, brown2025alphaearth}. While the relevance of spatial and temporal diversity of the pretraining data is subject to current research~\cite{purohit2025doesspatialdistributionpretraining}, one thing is clear: \emph{no GFM has seen it all, especially as landscapes keep constantly changing.}
Consequently, the first signal we want to analyze and use to assess the model's reliability is the input. However, the raw pixel space is both too high-dimensional and not very informative. As an interpretable alternative, we consider geophysical features of the scene, such as land cover and elevation or slope. These features can be seen as interpretable meta-features of the scene that could affect how certain high-stakes events, such as landslides, appear in RS imagery. 

\setlength{\tabcolsep}{5pt} 

\begin{table}[h]
\centering
\caption{Geophysical features from external datasets used to derive interpretable physical covariates, aligned with satellite imagery.}
\label{tab:physical_features}
\scalebox{0.9}{
\begin{tabular}{@{}lll@{}}
\toprule
\textbf{Dataset} & \textbf{Resolution} & \textbf{Included Features} \\ \midrule
HydroATLAS \cite{Linke2019} & 500m & \begin{tabular}[c]{@{}l@{}}Hydro-environmental \& \\  
land cover attributes\end{tabular}\\
\midrule
Copernicus DEM~\cite{CopernicusDEM2026} & 30m & \begin{tabular}[c]{@{}l@{}}Elevation\end{tabular} \\ \bottomrule
\end{tabular}
}
\end{table}

\begin{figure*}[h]
    \centering
    \includegraphics[width=\linewidth]{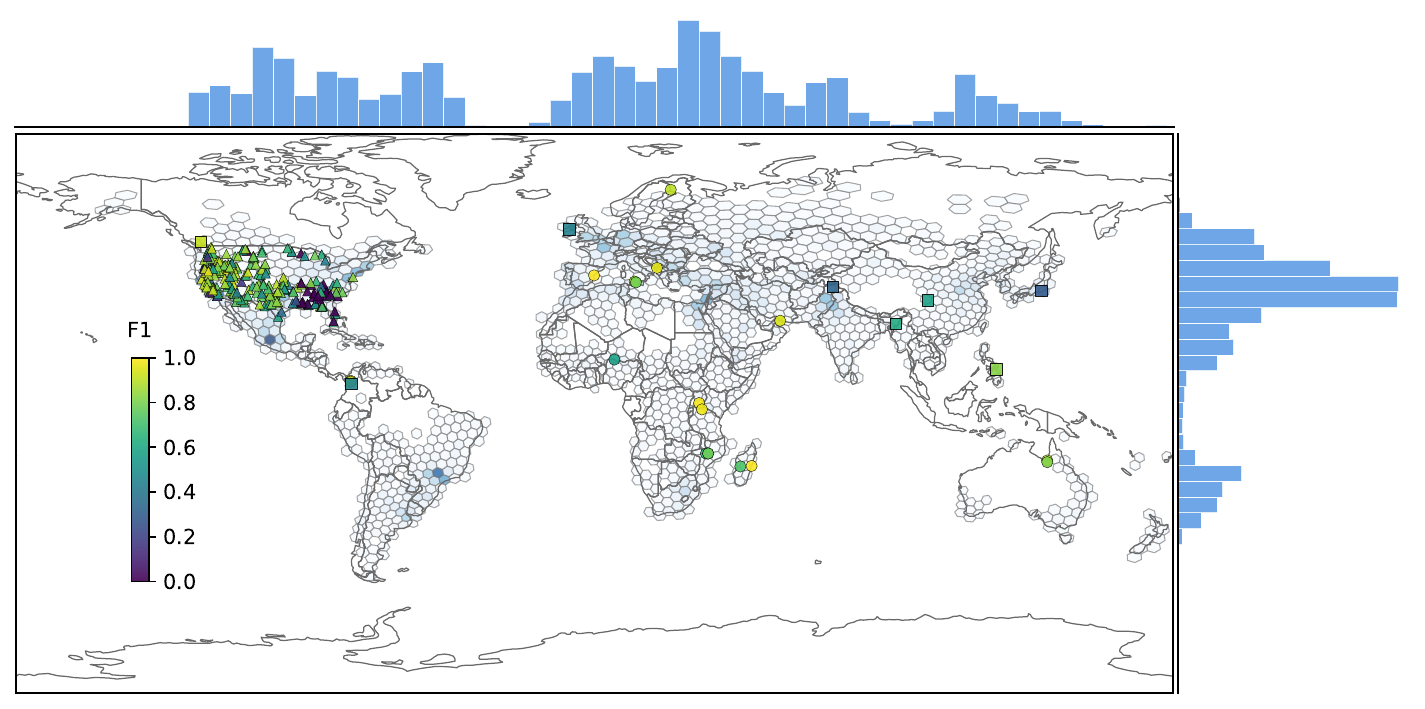}
    \caption{\textbf{Data distributions and performance.} F1 performance on the test set of the selected downstream tasks, overlaid on a hexagonal spatial distribution of SSL4EO-S12 across the globe. Shapes identify the task: $\triangle$ - burn scars, $\bigcirc$ - floods and $\square$ - landslides. }
    \label{fig:world_map}
\end{figure*}

The best choice of features to incorporate depends highly on the downstream task. To showcase the potential and task-agnostic flexibility of \idk{}, we choose HydroATLAS~\cite{Linke2019} for a wide range of hydro-environmental and land-cover attributes and Copernicus Digital Elevation Model~\cite{CopernicusDEM2026} for high-resolution elevation information. We match features from pretraining and downstream data by geographic location. 
For each feature $\phi_j$, we estimate the empirical CDF $F_j$ in the pretraining data and then compute the percentile rank of a new point $x$ of the downstream task:
\scorebox{%
\begin{equation}
\mathcal{U}_I^{Q_j}(x)=F_j(\phi_j(x)).
\end{equation}
}
These percentiles give a measure of feature extremity (the highest and lowest percentiles) of a sample relative to the pretraining data. Additionally, we estimate a spatial density map of the pretraining data (see hexagonal overlay on Supplementary Material (SM) Fig.\ref{fig:world_map}) over the globe and integrate the information into the downstream data samples. 
\scorebox{
\begin{equation}
\mathcal{U}_I^{\mathrm{density}}(x)= \hat{\rho}(s(x)),
\end{equation}
}
where $\hat{\rho}$ is the kernel density estimation fitted on pretraining locations and $s$ denotes the location of sample $x$.
Low density values indicate that there are few pretraining samples from nearby regions.



\paragraph{$\mathcal{U}_E$ -- The embedding signal}
\begin{quote}
\emph{I don’t really understand what I’m looking at…}
\hspace*{\fill} - \idk{}
\end{quote}

\begin{figure}
    \centering
    \includegraphics[width=\linewidth]{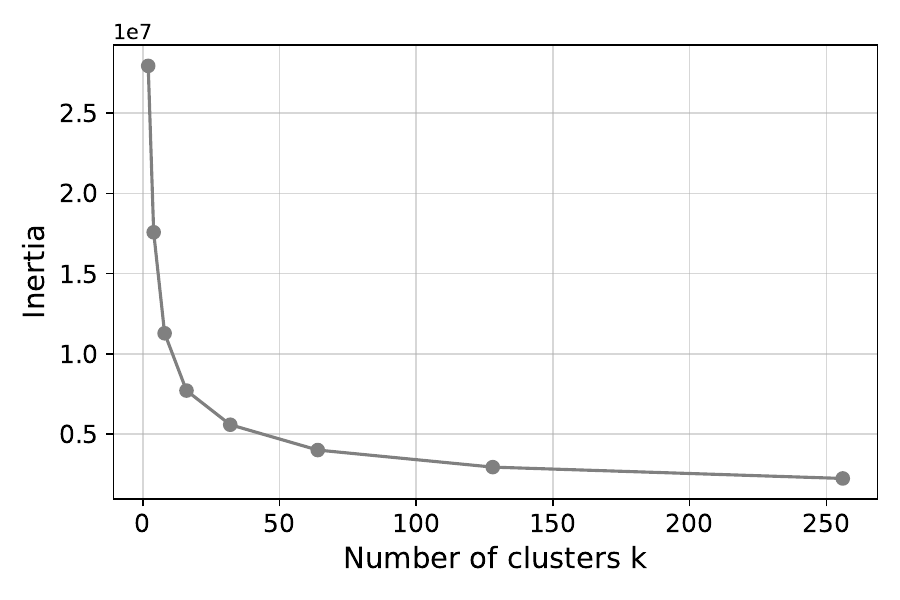}
    \caption{\textbf{Prototype selection for embedding familiarity.} $k$-means clustering of pretraining embeddings. The elbow criterion indicates stable performance around $k=64$.}
    \label{fig:kmeans_elbow}
\end{figure}

For the second reliability signal, we focus on the model's internal representation. Even if an input appears visually normal and all geophysical features seem regular, the embedding space can reveal when the model is ``confused'' by samples it has not encountered during pretraining. It could, for instance, be a known location that, due to land-use change or a disaster, becomes unfamiliar to the model in certain respects. 
Ideally, one would simply compute an embedding for a new sample and analyze its location relative to the pretraining data, for instance by retrieving its k-nearest neighbors and comparing their density to the overall density of the pretraining data. This, however, would require users to download and store the embeddings for the entire pretraining process, which, in some cases, are not publicly available and entail a substantial computational burden.
As a feasible alternative, we quantify the signal by clustering the pretraining embeddings into a set of representative prototypes that are cheap to store and provide. While we experimented with various clustering techniques, we settled on $k$-means for its suitability, using an available implementation in the FAISS library that scales to large datasets.
We experiment with different values of $k$, and use the elbow method to choose the preferred number of clusters (Fig.~\ref{fig:kmeans_elbow}). We choose $k=64$ and note that the results are robust to adjacent cluster numbers.

We compute two distinct out-of-distribution signals that capture different notions of a sample's unfamiliarity in the embedding space. First, we compute the normalized distance of the embedding $e$ of sample $x$ to its closest cluster. 
\scorebox{
\begin{equation}
\mathcal{U}_E^{\mathrm{nd}}(x)
= \frac{\|e(x)-\mu_{c(x)}\|}{\bar d_{c(x)}},
\end{equation}
}
where $c(x)$ denotes the closest cluster, $\mu_{c(x)}$ that cluster's center and $\bar{d}_{c(x)}$ the average pretraining sample distance within that cluster.
As this captures only the relation to the assigned cluster, we additionally compute the Nearest Centroid Distance Deficit (NCDD), an emerging OOD score from medical imagery~\citep{pokhrel2024ncdd}. The NCDD measures the relative closeness of a point to its assigned cluster center compared to all other centroids. NCDD is defined as: 

\scorebox{
\begin{equation}
\mathcal{U}_E^{\mathrm{NCDD}}(x)=\sum_{j\neq c(x)} d_j(x)-(k-1)\,d_{c(x)}(x),
\end{equation}
}
where $d_j(x)$ denote the cluster-normalized distances.
\noindent

\paragraph{$\mathcal{U}_T$ -- The task signal}
\begin{quote}
\emph{I can’t answer this with the information I have.}\\
\hspace*{\fill} - \idk{}
\end{quote}

Even when a new input $x$ lies within the expected distribution of the GFM's internal representations, the model may still fail to perform the downstream task accurately. The task signal $\mathcal{U}_T$ indicates when confidence collapses, flagging that the model can't solve the task with the available information. This is achieved by quantifying the model's predictive uncertainty. 

Uncertainty estimation and quantification have been extensively investigated in deep learning \cite{gawlikowski2023survey}. In this work, we equip the underlying model with uncertainty quantification capabilities using two ensemble-based approaches: the standard ensemble and the bootstrap ensemble~\cite{Lakshminarayanan2017simple}. Ensembles provide reliable uncertainty estimates and can be readily implemented without requiring major modifications to the deterministic model. Then, we compute standard uncertainty metrics at the pixel level, which, in a second step, we will translate to image-level uncertainties as signals $\mathcal{U}_T$. 
Given an ensemble of $N$ members, each providing a probabilistic prediction $f_i(x) = p_i$ in classification for an input $x$, we compute three key uncertainty measures.

\begin{enumerate}
    \item \textbf{Predicted Probability} ($\bar{p}$): The average of the individual member predictions, which represents the ensemble's consensus on the most likely class and is used for the model's final prediction.
    \begin{align}
    \bar{p} = \frac{1}{N} \sum_{i=1}^{N} p_i
    \end{align}
    
    \item \textbf{Average Entropy} ($\bar{\mathcal{H}}(x)$): Measures the average uncertainty across ensemble members. A high value indicates high aleatoric uncertainty inherent in the data.    
    \begin{align} 
    \bar{\mathcal{H}}(x) = \frac{1}{N} \sum_{i=1}^{N} \mathcal{H}(p_i)
    \end{align}
    
    where the entropy of a single prediction $p$ is defined as:
    \begin{align}
    \mathcal{H}(p) = -\sum_{c=1}^{C} p_c \log(p_c).
    \end{align}
    
    \item \textbf{Mutual Information} ($\mathcal{I}$): Captures the epistemic uncertainty, or the uncertainty due to a lack of model consensus. It is the difference between the total predictive entropy and the average entropy of the members.
    \begin{align}
    \mathcal{I}(x) = \mathcal{H}(\bar{p}) - \frac{1}{N} \sum_{i=1}^{N} \mathcal{H}(p_i)
    \end{align}
\end{enumerate}

From an information-theoretic perspective, the total uncertainty in a model's prediction—represented by the \textbf{Predictive Entropy} $\mathcal{H}(\bar{p})$—can be decomposed into these two distinct components. This additive decomposition allows for the formal separation of uncertainty stemming from the model parameters (\textit{epistemic}) from the uncertainty inherent in the data itself (\textit{aleatoric}):
\begin{equation}
    \underbrace{\mathcal{H}(\bar{p})}_{\text{Total Uncertainty}} = \underbrace{\mathcal{I}}_{\text{Epistemic (Model)}} + \underbrace{\left[ \frac{1}{N} \sum_{i=1}^{N} \mathcal{H}(p_i) \right]}_{\text{Aleatoric (Data)}}.
\end{equation}
Consequently, we use both mutual information and average entropy as task-specific uncertainty scores ($\mathcal{U}_T$) within our selective prediction framework. Predictive probabilities and ensemble-based uncertainty scores are used only as relative signals for selection, and hence, we do not additionally calibrate the uncertainties.


\noindent

\paragraph{Image-level Uncertainty.} To obtain a single uncertainty value for an entire image, we aggregate each previous pixel-level metric over a specific region of interest (ROI). For a given pixel-level uncertainty metric $\mathcal{M}(x)$ (e.g., Average Entropy), the image-level uncertainty, also referred to as $\mathcal{U}_{T}$, is the average of each metric over the ROI.
\scorebox{
\begin{equation}
\mathcal{U}^{\mathcal{M}}_{T} = \frac{1}{|\text{ROI}|} \sum_{\mathbf{x} \in \text{ROI}} \mathcal{M}(\mathbf{x}), \qquad\text{for }\mathcal{M}=\mathcal{I, \bar{\mathcal{H}}}
\end{equation}
}

\noindent
where $|\text{ROI}|$ is the number of pixels within the region of interest. We define the ROI as the predicted event area, defined as the set of pixels where the mean predicted probability $\bar{p}$ is greater than or equal to a threshold. 

Uncertainties depend strongly on the task at hand; therefore, the threshold for defining the predicted event area can vary across tasks. We define a range of thresholds, from the standard choice of 0.5 (for binary classification) to a lower threshold of 0.05, and analyze their contribution to the selective classifier.

\paragraph{Selective Classifier.}
The ultimate goal of \idk{} is to enable a selective predictor $h$ that knows when to provide a trustworthy output and when to abstain. This requires a selection function $g$ that maps the fused signals $(\mathcal{U}_I, \mathcal{U}_E, \mathcal{U}_T)$ to a binary decision: accept or reject.

\paragraph{Defining Failure.} The utility of selective prediction hinges on a clear definition of what constitutes a ``success'' versus a ``failure''. This definition is inherently task-dependent and varies with the end-user's risk tolerance. For disaster response tasks such as burn scar mapping or flood detection, a ``failure'' might be defined by low precision (causing false alarms) or low recall (missed critical events). In this work, we aim to demonstrate the framework's broad applicability and to set a simple threshold based on the $F1$ score. Given the overall higher difficulty (in terms of average performance) of landslide detection, we adopt a lower acceptance threshold ($F1 \geq 0.5$) for this task, whereas all other tasks use $F1 \geq 0.6$.
This results in an overall failure rate of $0.33$ (burn scars), $0.21$ (floods), and $0.53$ (landslides) on the test sets.
Using this ground truth, we treat selective prediction as a secondary classification task: identifying scenes where the downstream model's performance is likely to fall below this safety margin. For the complete F1 performance per test set, see SM Fig. \ref{fig:world_map}. 

\paragraph{Feature Selection and Training.} Given the high dimensionality of the combined input, embedding, and task-level flags (e.g., 21 physical covariates from HydroATLAS and Copernicus DEM plus 4 UQ/OOD scores), we first discard highly correlated features and then use a heuristic to prune the input signals down to the 7 most informative ones. Afterward, we perform Recursive Feature Elimination (RFE) on the remaining input, along with the UQ and OOD scores, to identify the most discriminative signals for each downstream task. For each task, we then tune and train a shallow decision tree (\verb|max_depth=3|) to serve as the selection function $g$. We specifically choose a shallow, glass-box architecture to maintain interpretability for practitioners. This allows users to not only see when the model abstains but to understand why (e.g., identifying that a prediction was rejected due to a specific combination of high terrain slope and embedding drift). Detailed configurations for feature selection and hyperparameter tuning are provided in SM \ref{app:impl-selective-clf}.
To train and tune the classifier, we use only the training and validation splits from the downstream datasets. The final evaluation is then carried out on the test split.

We have also evaluated low-degree polynomial models and neural additive models, which yield comparable performance. 
We adopt decision trees for their transparency. 


\section{Data and Experimental Setup}
The \idk{} framework is model-agnostic, supporting diverse EO tasks and GFMs. In this work, we exemplify its potential by building on the SSL4EO-S12 suite of foundation models \cite{wang2023ssl4eo}, which are trained on a large-scale multi-seasonal Sentinel-1 and Sentinel-2 (S2) imagery covering ~250,000 locations worldwide. The S2 data includes 13 spectral bands (B1–B8A, B9, B11, B12), preprocessed to a common 10m resolution. The SSL4EO-S12 model suite includes pre-trained models covering different representative self-supervised learning methods, such as MAE~\cite{9879206}, DINO~\cite{9709990}, and MoCo~\cite{9157636}, and various established vision model architectures.
For our analysis, we choose a ViT/s 16 model with $\sim$23 million parameters, pretrained with the contrastive objective MoCo: a compact Vision Transformer designed to balance state-of-the-art performance and size for running comprehensive analyses. 

We demonstrate the framework's applicability by integrating \idk{} into the pipelines of the following risk-sensitive applications:

\begin{itemize}
    \item \textbf{Burn scar mapping with ExeBench \citep{zhao2025exebenchbenchmarkingfoundationmodels}}

    Harmonized satellite imagery from Landsat and Sentinel-2 \cite{claverie2018harmonized} collected over burn scar areas between $2018$ and $2021$ over the United States. It contains $804$ images of size $512 \times 512$ with $6$ spectral bands (visible, infrared, near-infrared, and shortwave infrared), divided into $485$, $54$, and $263$ for training, validation, and testing, respectively. The ground resolution is $30$ meters per pixel.
    
    \item \textbf{Flood mapping with WorldFloods \cite{PortalesJulia2023}}

    Compilation of 509 globally distributed Sentinel-2 Level 1C full-image scenes depicting flood events from 2016 to 2019. In line with prior work, we slice the scenes into smaller patches of $256 \times 256$ pixels, filtering out up to 10\% of the tolerable cloud coverage. The resulting data is split into 39060, 1854, and 1936 samples for training, validation, and testing, respectively.

    \item \textbf{Landslide detection with Event Reference Data \cite{Orynbaikyzy2025}}

    Compilation of 28 single events of reported landslides with varying extent that occurred worldwide from July 2015 onwards. For the same landslide event, we work with the Sentinel-2 bands: visible and near-infrared bands with 10m spatial resolution, and four red-edge bands and two short-wavelength near-infrared bands with 20m resolution. The large Sentinel-2 scenes are slices into smaller patches of $64 \times 64$ pixels, resulting in 225, 41 and 68 samples for training, validation and testing, respectively.
    
\end{itemize}

As a decoder for the downstream task, we chose a common deep CNN with $\sim 41$ million parameters. By choosing an expressive decoder, we make sure that the decoder is not the bottleneck for performance issues and failures can be linked to the encoder FM or the data.
For all tasks, patches are resized to $224 \times 224$ pixels to match the SSL4EO pretraining sample format. 
We evaluate both standard ensembles and bootstrap ensembles and select the variant with the best validation performance. This results in bootstrap ensembles for burn scar mapping and standard ensembles for flood and landslide detection.

\paragraph{Evaluation Metrics.}
To rigorously assess the utility of the derived flags, we use metrics that quantify the system's performance across varying rejection settings.
\begin{itemize}
\item Risk-Coverage (RC) Curves: We plot the Risk (failure rate among accepted samples) on the y-axis against Coverage (the fraction of samples kept) on the x-axis.
\item Area Under the Risk-Coverage Curve (AURC): We use the AURC as a scalar summary of the RC trade-off, where a lower value indicates a more efficient prioritization of failures for rejection.
\item Discard-Performance (DP) Curves: We plot the average performance in F1 over the accepted samples on the y-axis against Coverage (the fraction of samples kept) on the x-axis.
\item Area Under the Discard-Performance Curve (AUDP): We use the AUDP as a scalar summary of the DP trade-off, where a higher value indicates more efficient prioritization of low-performing outcomes for rejection.
\item Operational Risk: We report risk at specific, practically relevant coverage levels (e.g., Risk@0.5, 0.75) to guide real-world deployment, where a certain data volume must be maintained.
\end{itemize}

\paragraph{Curve Estimation.}
Due to the discrete outputs of the decision tree, multiple samples may share the same rejection score. To avoid arbitrary ordering effects when computing risk–coverage and discard–performance curves and metrics, we randomly permute the dataset 100 times and average the resulting curves. Final uncertainty bands reflect variability across 30 independent training seeds (which affects both the automatic feature selection and cross-fold hyperparameter tuning), after averaging the curves obtained from the score permutations.

\paragraph{Baselines.}
For comparison, we compute the same metrics using either the single uncertainty score or the out-of-distribution score, as is commonly done. Additionally, we include a random discard baseline, where we discard the test images in random order. We run the random baseline 100 times and report its mean and $95\%$ confidence bands.  
\section{Results and Discussion}
\subsection{Selective Prediction Performance}
\begin{figure*}
    \centering
    \includegraphics[width=0.97\linewidth]{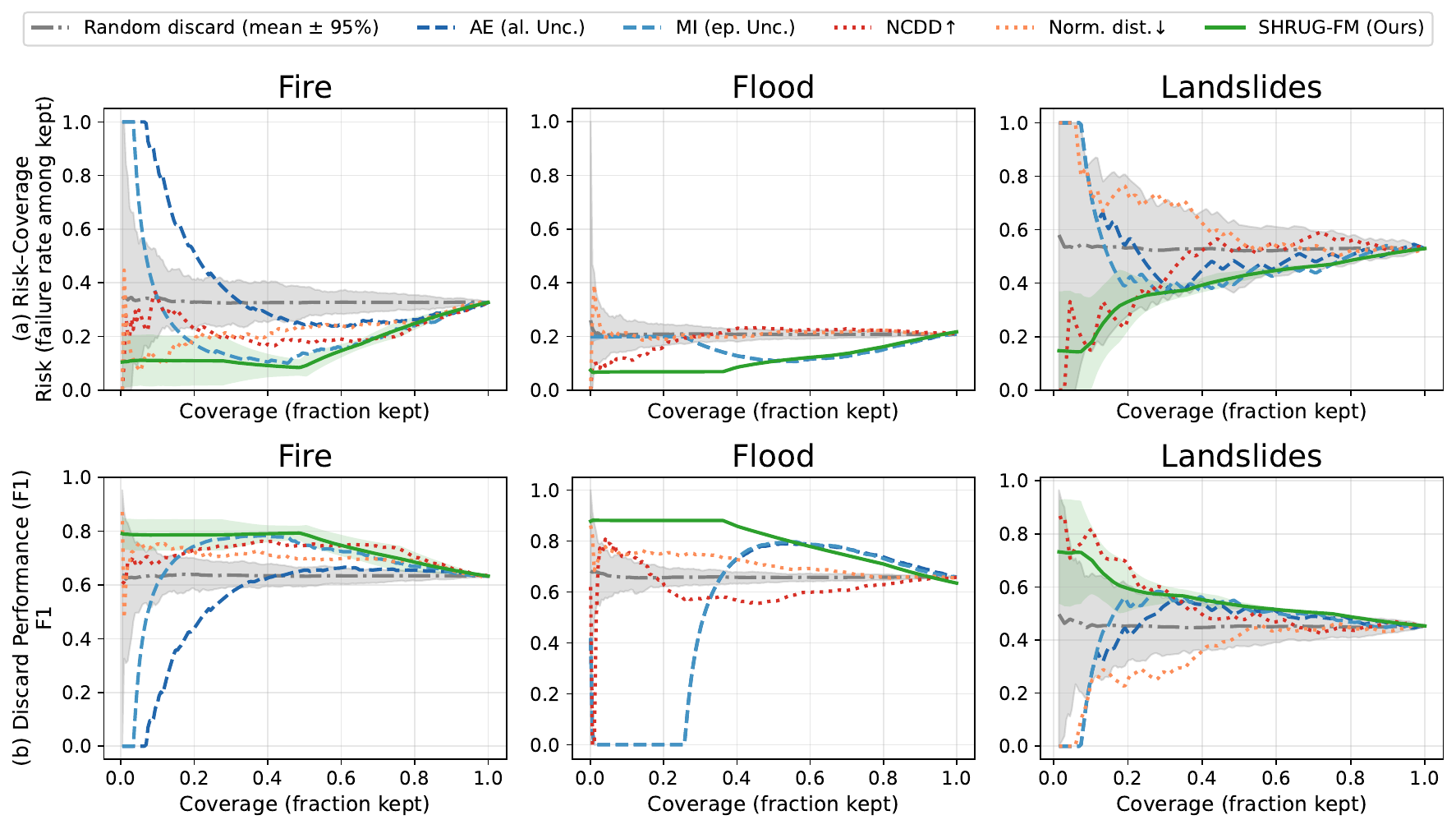}
    \caption{\textbf{Risk-Coverage and Discard-Performance Curves.} \textbf{(a - first row)}: Selective prediction performance by task. Risk–coverage curves for each task. Lower curves indicate more effective identification and rejection of failure cases. \idk{} consistently achieves lower risk at comparable coverage than individual uncertainty or OOD signals. Shaded regions indicate $95\%$ confidence bands. \textbf{(b - second row)}: Performance under selective abstention. Discard–performance curves showing average F1 over accepted samples as coverage varies. Higher curves indicate more effective prioritization of reliable predictions.}
    \label{fig:risk_discard_performances_curves}
\end{figure*}

The primary objective of \idk{} is to provide a reliable ``abstain'' mechanism for geospatial foundation models. As demonstrated in Fig.~\ref{fig:risk_discard_performances_curves} (a), our fused selective classifier consistently outperforms all single-signal baselines across the three disaster response tasks.

\textbf{Risk Reduction.} In the Fire task, \idk{} achieves a Risk@0.5 of $0.090 \pm 0.038$, improving substantially over the best single-signal baseline (Mutual Information at $0.137$). On WorldFloods, the framework achieves the lowest overall Area Under the Risk-Coverage Curve (AURC) of $0.115 \pm 0.001$ and yields competitive risk reduction for uncertainty scores at risk levels above $0.5$. For Landslides, despite extreme data scarcity ($N=225$ in training), our framework successfully identifies a high-confidence subset, reducing the AURC to $0.392 \pm 0.035$ compared to $0.526$ with Aleatoric Uncertainty alone.

\textbf{Discard-Performance Curves.} The F1-discard plots (Fig.~\ref{fig:risk_discard_performances_curves}) (b) re-emphasize \idk{}'s robustness. Single-signal baselines such as Entropy or MI quickly level off, suggesting a ``healthy amount of uncertainty'' regime. At this level, we tend to discard actually well-performing samples. Overall, \idk{} at least maintains or improves performance at higher coverage.

\subsection{Ablation Study and the ``Input Paradox''}
The ablation analysis (Table~\ref{tab:ablation}) reveals a critical dependency on Uncertainty Quantification (UQ) signals. Removing UQ metrics leads to a catastrophic collapse in Flood ROC AUC from $0.725$ to $0.524$. 
However, we observe an ``Input Paradox'' in which the ``No Input Feature'' configuration, in which we omit the input signals, occasionally yields a higher ROC AUC than the ``All'' configuration (e.g., $0.84$ vs $0.791$ for Fire). This suggests that, while geophysical features are highly interpretable, they may introduce noise in low-data regimes, complicating the optimization of a shallow tree. Nevertheless, we retain these features in our primary framework as they provide the physical grounding necessary for human-in-the-loop disaster response.

\subsection{Interpretability and Feature Persistence}
\begin{figure*}[h]
    \centering
    \includegraphics[width=\linewidth]{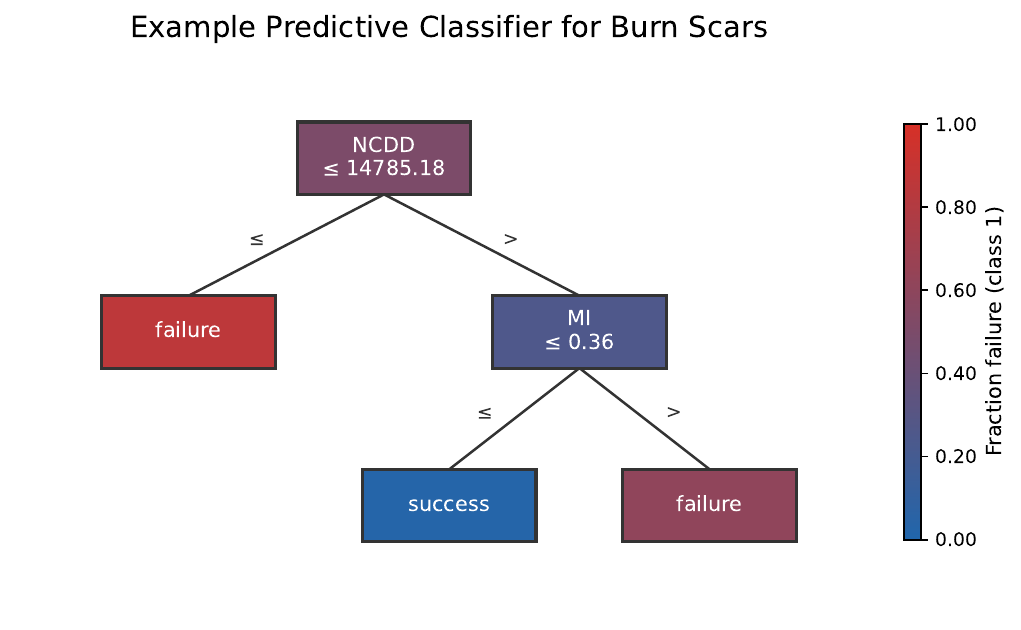}
    \caption{\textbf{Example selective classifier.} Shallow decision tree learned from reliability signals on the burn scars task. The model uses NCDD as a first signal. When it lies below a threshold which indicates a stronger out of distribution signal, it classifies as failure. Above the threshold, mutual information corresponding to epistemic uncertainty is used for another branching-off with high uncertainty indicating failure and low uncertainty success. The structure reflects intuitive failure modes, combining complementary signals from the scores.}
    \label{fig:tree_plot}
\end{figure*}

A key benefit of \idk{} is its transparency. An example decision tree for Burn Scars (see Fig.~\ref{fig:tree_plot}) identifies thresholds for NCDD scores and epistemic uncertainty (Mutual Information) where a low NCDD and high uncertainty both indicate failure.

\textbf{Feature Stability.} Across 30 independent runs, \textbf{Mutual Information} was selected in $100\%$ of the trees for every task, confirming that model disagreement is the most robust universal indicator of GFM failure. In the Flood task, \textbf{spatial density} also achieved $100\%$ persistence, demonstrating that the selective classifier effectively flags regions far from the pretraining data distribution. In Landslides, the \textbf{NCDD} and \textbf{normalized distance} achieved $70\%$ and $100\%$ persistence, respectively, underscoring the need for embedding-space OOD detection for failure-prone events.

Qualitatively, task-level epistemic uncertainty (mutual information) most reliably identifies model disagreement and reduces the number of high-risk samples for the fire task. Spatial density is particularly informative for floods, where geographic coverage gaps may affect failure rates (see SM \cref{fig:world_map}). Embedding OOD (NCDD / normalized distance) seem to be most useful for rare, structurally atypical events (landslides). These patterns confirm that the three signals capture different failure causes.



\setlength{\tabcolsep}{3pt} 
\begin{table*}[htbp]
\scriptsize
\centering
\caption{Risk–coverage: AUC and risk at coverage levels. Aggregate selective prediction metrics. Area under the risk–coverage curve and risk at two coverage levels (0.5, 0.75) (for all metrics, lower is better). Results reported on held-out test sets. \idk{} consistently improves overall AUC and shows at least competitive reliability at fixed coverage levels.}
\label{tab:risk_coverage_combined}
\begin{tabular}{lccc|ccc|ccc}
\toprule
  & \multicolumn{3}{c}{Fire} & \multicolumn{3}{c}{Flood} & \multicolumn{3}{c}{Landslides} \\
\midrule
   &   & \multicolumn{2}{c}{Risk@} &   & \multicolumn{2}{c}{Risk@} &   & \multicolumn{2}{c}{Risk@} \\
Score & AUC$\downarrow$ & 0.5$\downarrow$ & 0.75$\downarrow$ & AUC$\downarrow$ & 0.5$\downarrow$ & 0.75$\downarrow$ & AUC$\downarrow$ & 0.5$\downarrow$ & 0.75$\downarrow$ \\
\midrule
AE (al. U.) & 0.391 $\pm$ 0.000 & 0.244 $\pm$ 0.000 & 0.259 $\pm$ 0.000 & 0.160 $\pm$ 0.001 & 0.110 $\pm$ 0.000 & 0.142 $\pm$ 0.000 & 0.526 $\pm$ 0.000 & 0.471 $\pm$ 0.000 & 0.508 $\pm$ 0.000 \\
MI (ep. U.) & 0.240 $\pm$ 0.000 & 0.137 $\pm$ 0.000 & 0.228 $\pm$ 0.000 & 0.159 $\pm$ 0.001 & 0.108 $\pm$ 0.000 & \textbf{0.139 $\pm$ 0.000} & 0.491 $\pm$ 0.000 & \textbf{0.412 $\pm$ 0.000} & \textbf{0.471 $\pm$ 0.000} \\
NCDD↑ & 0.225 $\pm$ 0.000 & 0.183 $\pm$ 0.000 & \textbf{0.203 $\pm$ 0.000} & 0.198 $\pm$ 0.000 & 0.228 $\pm$ 0.000 & 0.223 $\pm$ 0.000 & 0.451 $\pm$ 0.000 & 0.529 $\pm$ 0.000 & 0.569 $\pm$ 0.000 \\
Norm. dist.↓ & 0.224 $\pm$ 0.000 & 0.236 $\pm$ 0.000 & 0.253 $\pm$ 0.000 & 0.210 $\pm$ 0.000 & 0.211 $\pm$ 0.000 & 0.219 $\pm$ 0.000 & 0.614 $\pm$ 0.000 & 0.559 $\pm$ 0.000 & 0.510 $\pm$ 0.000 \\
\idk{} (Ours) & \textbf{0.160 $\pm$ 0.045} & \textbf{0.090 $\pm$ 0.038} & 0.223 $\pm$ 0.029 & \textbf{0.115 $\pm$ 0.001} & \textbf{0.107 $\pm$ 0.004} & 0.149 $\pm$ 0.000 & \textbf{0.392 $\pm$ 0.035} & 0.425 $\pm$ 0.021 & 0.471 $\pm$ 0.000 \\
\bottomrule
\end{tabular}
\end{table*}




\paragraph{Limitations.}
\idk{} aims to bridge the gap between benchmark performance and real-world use, driving the integration of novel GFMs into risk-sensitive applications while maintaining reliable, trustworthy outputs. This flexibility requires several task-dependent design choices, including the definition of failure criteria, the selection of physical input features, and the configuration of the selective classifier. While these decisions enable adaptation to diverse applications, they introduce sensitivity to hyperparameter choices, particularly in low-data regimes. Additionally, some signals require auxiliary information about the FM pretraining distribution. The input signal depends on access to the FM’s pretraining data and attribute alignments, which can be memory-intensive and are not always provided by GFM developers. Similarly, even when operating in a reduced-dimensional space such as the embedding space, clustering computations for the embedding signal may still be constrained by its size. We view this as an opportunity to enable closer integration between FM development and deployment, in which such metadata could be distributed alongside pretrained models.
Although ensembles are straightforward to implement, we are also aware of their high computational demands. This could be mitigated by integrating more lightweight uncertainty quantification methods. We believe that density-predicting models could be a suitable alternative that yields a full predictive distribution in a single forward pass.


\paragraph{Deployment considerations.}
In operational deployments, \idk{} would be used as a front-line triage: predictions flagged by $g$ are routed to human analysts (or higher-fidelity models), whereas accepted predictions are routed to automated pipelines. We recommend (i) exposing the top contributing signals per abstention to the reviewer, (ii) periodic recalibration using newly labelled data, and (iii) distributing precomputed embedding prototypes and pretraining feature quantiles alongside FMs to reduce integration friction.

The proposed signals are designed to capture complementary aspects of reliability, but they are not statistically independent. Distributional extremeness in physical input space may propagate to embedding unfamiliarity and elevated predictive uncertainty. While this redundancy can improve robustness when individual signals fail, it also highlights that reliability estimation remains sensitive to distribution shifts in downstream data. In practical deployments, periodic recalibration or updates may therefore be required. 


\section{Conclusions}
\idk{} offers the opportunity to accelerate the safe deployment of GFM by converting uncertainty into actionable signals, fostering trust and accountability in AI deployment for environmental management. Our quantitative results establish that the presented selective mechanism effectively improves reliability by reducing the prediction risk (failure rate) on retained samples as the coverage decreases, isolating high-confidence predictions in an interpretable way. The present study evaluates a representative subset of foundation models, downstream tasks, and model configurations. The space of possible combinations is substantially broader, and future work is needed to assess generalization across additional architectures, training paradigms, and application domains. We also aim to collaborate with domain experts to translate these combined flags into actionable, interpretable diagnostics. 
This can guide future GFM pretraining data strategies and potentially yield new scientific insights into the EO data itself.

\section*{Acknowledgments}
This work is a research product of ESL (\href{http://eslab.ai/}{ESL.ai}), an initiative of the Frontier Development Lab, delivered by Trillium Technologies in partnership with ESA, University of Oxford, Google Cloud, and NVIDIA. M Gonzalez-Calabuig thanks the Horizon project ELIAS (grant agreement 101120237). K H Cohrs acknowledges the support from the European Research Council (ERC) under the ERC Synergy Grant USMILE (grant agreement 855187). V Nedungadi acknowledges support from the Digital Europe Programme under Grant agreement \href{http://www.agrifoodtef.eu}{AgrifoodTEF} - Test and Experiment Facilities for the Agri-Food Domain (Grant \#\href{https://ec.europa.eu/info/funding-tenders/opportunities/portal/screen/opportunities/projects-details/43152860/101100622}{101100622}). V Sitokonstantinou acknowledges the support of the project `WUR FM' (project number KB-57-001-002).


{
    \small
    \bibliographystyle{ieeenat_fullname}
    \bibliography{main}
}
\clearpage
\setcounter{page}{1}
\maketitlesupplementary

\section{Appendix: Additional Results}
We provide a set of ablations regarding the importance and usage of the different reliability signals in \cref{tab:ablation} and \cref{tab:feature_stability}.

\begin{table*}
\footnotesize
\centering
\begin{tabular}{lccc|ccc}
\toprule
 & \multicolumn{3}{c}{Classifier ROC AUC$\uparrow$} & \multicolumn{3}{c}{Risk--Coverage AUC$\downarrow$} \\
\cmidrule(lr){2-4} \cmidrule(lr){5-7}
 & Fire & Flood & Landslides & Fire & Flood & Landslides \\
\midrule
All & 0.791 $\pm$ 0.0635 & 0.725 $\pm$ 0.00378 & 0.654 $\pm$ 0.0263 & 0.16 $\pm$ 0.0435 & 0.114 $\pm$ 0.00124 & 0.392 $\pm$ 0.0329 \\
No Input Feature Signals & \textbf{0.84 $\pm$ 0.00944} & \textbf{0.739 $\pm$ 0.00106} & \textbf{0.661 $\pm$ 0.0382} & \textbf{0.13 $\pm$ 0.0105} & 0.104 $\pm$ 0.000206 & \textbf{0.381 $\pm$ 0.0385} \\
No Embedding OOD Signals & 0.77 $\pm$ 0.0224 & 0.724 $\pm$ 0.00365 & 0.622 $\pm$ 0.0373 & 0.173 $\pm$ 0.0258 & 0.114 $\pm$ 0.0012 & 0.423 $\pm$ 0.0823 \\
No UQ (MI, AE) Signals & 0.706 $\pm$ 0.0242 & 0.524 $\pm$ 0.024 & 0.548 $\pm$ 0.0248 & 0.227 $\pm$ 0.0282 & 0.207 $\pm$ 0.00605 & 0.499 $\pm$ 0.00608 \\
UQ only & 0.8 $\pm$ 0.00788 & 0.739 $\pm$ 0.000477 & 0.633 $\pm$ 0.0104 & 0.138 $\pm$ 0.0102 & \textbf{0.104 $\pm$ 0.000229} & 0.412 $\pm$ 0.039 \\
Dist only & 0.739 $\pm$ 0.0465 & 0.495 $\pm$ 0.00921 & 0.562 $\pm$ 0.0181 & 0.198 $\pm$ 0.0382 & 0.21 $\pm$ 0.00418 & 0.494 $\pm$ 0.00732 \\
Input only & 0.473 $\pm$ 0.0556 & 0.533 $\pm$ 0.0056 & 0.417 $\pm$ 0.0529 & 0.351 $\pm$ 0.0404 & 0.205 $\pm$ 0.00193 & 0.604 $\pm$ 0.0495 \\
\bottomrule
\end{tabular}
\caption{Ablation of reliability signals. Selective prediction performance using input-only, embedding-only, task-only, and combined signals. For each scenario, we withheld a subset of the signals for the whole selective prediction training pipeline.}
\label{tab:ablation}
\end{table*}

\begin{table*}
\centering
\begin{tabular}{llll}
\toprule
 & Fire & Flood & Landslide \\
feature &  &  &  \\
\midrule
\verb|Average_Entropy (0.05)| & --- & 100.0\% & 16.7\% \\
\verb|Average_Entropy (0.3)| & 90.0\% & --- & --- \\
\verb|Mutual Information (0.05)| & --- & 100.0\% & 100.0\% \\
\verb|Mutual Information (0.1)| & 100.0\% & --- & --- \\
\verb|density| & 53.3\% & 100.0\% & --- \\
\verb|ncdd| & 30.0\% & --- & 70.0\% \\
\verb|normalized_distance| & --- & --- & 100.0\% \\
\verb|ria_ha_ssu| & --- & --- & 3.3\% \\
\bottomrule
\end{tabular}
\caption{Signals used by the selective classifier. Most informative reliability indicators were identified via recursive feature elimination and feature importance of the resulting tree. Decimals after the uncertainty scores denote the threshold on the predicted probability, marking the region of interest over which the measure is averaged for the image-level uncertainty. }
\label{tab:feature_stability}
\end{table*}

\section{Appendix: Implementation Details for Selective Classifier}
\label{app:impl-selective-clf}

In the very first step, we remove highly correlated physical features. Next, feature subsets are selected using RFECV with decision trees over all candidate signals $(\mathcal{U}_I,\mathcal{U}_E,\mathcal{U}_T)$.

For model selection, we tune decision tree hyperparameters using cross-validation. Fire and landslide tasks use repeated, stratified $k$-fold validation, while flood detection uses standard $k$-fold validation. The maximum tree depth is limited to three in all experiments.

Final models are trained on the combined training and validation splits. Reported performance reflects evaluation on the held-out test set.

\section{Appendix: Feature Selection and Stability Analysis}

To ensure the robustness of the \idk{} framework, we repeat the entire pipeline—including RFE-based feature selection and tree tuning—across 30 independent runs with varying random seeds. We observe minimal sensitivity to the seed, indicating that our framework successfully identifies stable, physically-grounded failure patterns.

\begin{table*}[ht]
\centering
\caption{Full set of input features for the selective classifier $g$.}
\label{tab:feature_set}
\small
\begin{tabular}{lll}
\toprule
\textbf{Category} & \textbf{Feature} & \textbf{Description} \\
\midrule
\textbf{HydroATLAS} & \texttt{inu\_pc\_smn} & Min. annual inundation extent (\% cover). \\
 & \texttt{inu\_pc\_smx} & Max. annual inundation extent (\% cover). \\
 & \texttt{ria\_ha\_ssu} & Total river area (hectares). \\
 & \texttt{slp\_dg\_sav} & Average terrain slope (degrees $\times$ 10). \\
 & \texttt{snw\_pc\_syr} & Avg. annual snow cover extent (\% cover). \\
 & \texttt{snw\_pc\_smx} & Max. annual snow cover extent (\% cover). \\
 & \texttt{glc\_cl\_smj} & Land cover classes (spatial majority). \\
 & \texttt{wet\_pc\_sg1} & Wetland extent (grouping 1, \% cover). \\
 & \texttt{wet\_pc\_sg2} & Wetland extent (grouping 2, \% cover). \\
 & \texttt{for\_pc\_sse} & Average forest cover extent (\% cover). \\
 & \texttt{crp\_pc\_sse} & Average cropland extent (\% cover). \\
 & \texttt{pst\_pc\_sse} & Average pasture extent (\% cover). \\
 & \texttt{ire\_pc\_sse} & Average irrigated area extent (\% cover). \\
 & \texttt{urb\_pc\_sse} & Average urban extent (\% cover). \\
 & \texttt{hft\_ix\_s09} & Human Footprint index (year 2009). \\
\midrule
\textbf{DEM} & \texttt{elev\_mean} & Average elevation (meters a.s.l.). \\
 & \texttt{elev\_min} & Minimum elevation (meters a.s.l.). \\
 & \texttt{elev\_max} & Maximum elevation (meters a.s.l.). \\
\midrule
\textbf{Spatial} & \texttt{density} & Spatial density of pre-training data. \\
\midrule
\textbf{Embedding} & \texttt{NCDD} & Centroid Distance Deficit OOD score. \\
 & \texttt{norm\_dist} & Distance to closest pre-training cluster. \\
\midrule
\textbf{Uncertainty} & \texttt{avg\_entropy} & Image-level average aleatoric uncertainty. \\
 & \texttt{mutual\_info} & Image-level average epistemic uncertainty. \\
\bottomrule
\end{tabular}
\end{table*}

\end{document}